\pdfoutput=1

\documentclass[11pt]{article}

\usepackage[final]{acl}

\usepackage{times}
\usepackage{latexsym}
\usepackage{natbib}
\usepackage{amsmath,amssymb,amsfonts}
\usepackage{graphicx}
\usepackage{booktabs}
\usepackage{multirow}
\usepackage{caption} 
\usepackage{subcaption}
\usepackage[T1]{fontenc}
\usepackage[utf8]{inputenc}
\usepackage{microtype}
\usepackage{inconsolata}
\usepackage{xspace}

\newcommand{\methodnosapce}{MindRef}
\newcommand{\method}{MindRef\xspace}

\title{MindRef: Mimicking Human Memory for Hierarchical Reference Retrieval with Fine-Grained Location Awareness}

\author{
Ye Wang\textsuperscript{1}$^\text{*}$, 
Xinrun Xu\textsuperscript{2,3}$^\text{*}$, 
Zhiming Ding\textsuperscript{3}$^\text{\dag}$\\
\textsuperscript{1}{Renmin University of China}\\
\textsuperscript{2}{University of Chinese Academy of Sciences}\\ \textsuperscript{3}{Institute of Software, Chinese Academy of Sciences}\\
\texttt{yewang@ruc.edu.cn, xuxinrun20@mails.ucas.ac.cn, zhiming@iscas.ac.cn}
}

\begin{document}
\maketitle

\renewcommand{\thefootnote}{\fnsymbol{footnote}}
\footnotetext[1]{Equal Contribution.} 
\footnotetext[2]{Corresponding Author.}
\renewcommand{\thefootnote}{\arabic{footnote}}

\begin{abstract}
When completing knowledge-intensive tasks, humans sometimes need an answer and a corresponding reference passage for auxiliary reading. 
Previous methods required obtaining pre-segmented article chunks through additional retrieval models. 
This paper explores leveraging the parameterized knowledge stored during the pre-training phase of large language models (LLMs) to recall reference passage from any starting position independently. 
We propose a two-stage framework that simulates the scenario of humans recalling easily forgotten references. 
Initially, the LLM is prompted to recall document title identifiers to obtain a coarse-grained document set. 
Then, based on the acquired coarse-grained document set, it recalls fine-grained passage. 
In the two-stage recall process, we use constrained decoding to ensure that content outside of the stored documents is not generated. 
To increase speed, we only recall a short prefix in the second stage, and then locate its position to retrieve a complete passage. 
Experiments on KILT knowledge-sensitive tasks have verified that LLMs can independently recall reference passage locations in various task forms, and the obtained reference significantly assists downstream tasks.
\footnote{Code is available at \url{https://github.com/www-Ye/MindRef}.}
\end{abstract}

\section{Introduction}

Knowledge-intensive tasks rely heavily on large knowledge sources \citep{petroni-etal-2021-kilt}. 
Traditional methods often use retrieval models to find relevant passages from resources like Wikipedia for tasks such as question answering \cite{izacard-grave-2021-leveraging}. 
However, limitations exist with both sparse (lack of semantic depth) and dense retrieval (limited interaction between question and passage representations) \cite{khattab-etal-2021-relevance}.
Generative retrieval methods, leveraging models' generative abilities for deeper interaction with knowledge sources, are gaining popularity \cite{tay2022transformer}. 
However, Current retrieval methods require pre-segmented passages, limiting reference flexibility like human memory. 
We ask: "Can LLMs bypass chunking to recall references from any position?"

\begin{figure}[!ht]
\centering
\includegraphics[width=0.9\linewidth]{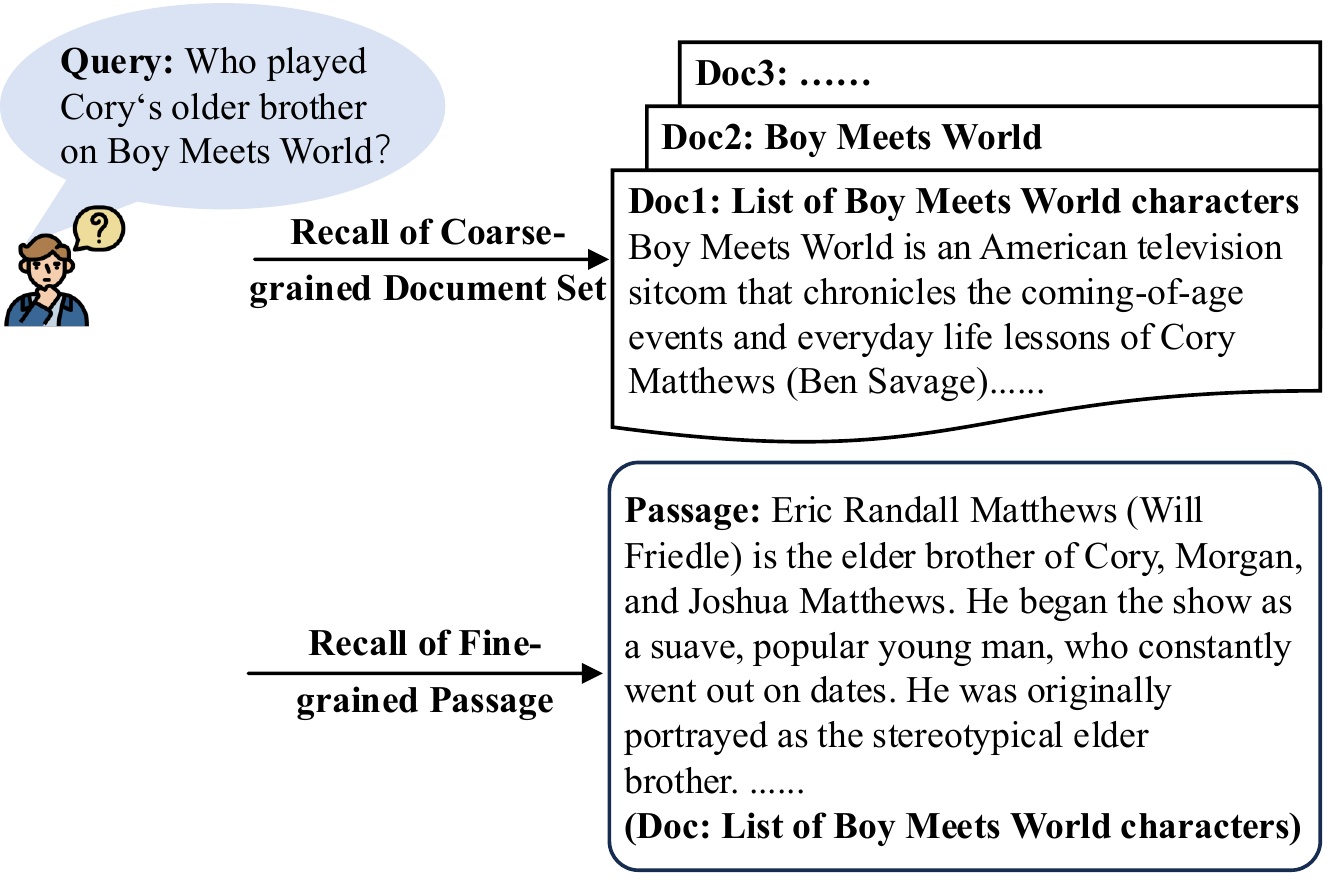} 
\caption{Human recall of forgotten information often involves a two-step process: recalling memorable documents first, then locating the specific passage within.}
\label{fig1}
\end{figure}

Leveraging the capabilities of LLMs, we propose \textbf{\methodnosapce}, a two-stage framework for flexible passage retrieval. 
Inspired by human recall, we first prompt the LLM to recall relevant document titles, guided by a Trie \cite{cormen2022introduction}. 
Then, using an FM-index \cite{ferragina2000opportunistic} built from the retrieved documents, the LLM recalls specific passages with flexible starting points. 
A weighted score combines both stages for final reference selection.
To enhance efficiency, \method retrieves passages by recalling only a short prefix. 
The LLM generates this prefix, which is located within the documents using the FM-index and KMP algorithm. 
Ultimately, the algorithm identifies a longer passage as the final reference.
This approach allows LLMs to access and retrieve natural references from articles of any length without relying on additional retrieval models or pre-segmentation, offering both flexibility and efficiency.

Extensive experiments on 6 KILT benchmark tasks \cite{petroni-etal-2021-kilt} demonstrate the effectiveness of \textbf{\methodnosapce}, enabling open-source LLMs like LLaMA \cite{touvron2023Llama} and LLaMA-2 \cite{touvron2023Llama2} to retrieve documents and passages, improving downstream task performance effectively. Key contributions include:

1) We propose \method, a cognitively-aligned retrieval framework that formalizes the `document-to-detail' mechanism of human memory into neural architectures for the first time. 
2) Breaking away from pre-chunking paradigms, our method achieves chunkless reference localization through joint Trie-FMIndex constrained decoding, enabling retrieval from arbitrary document positions. 
3) The SPRL co-optimization strategy delivers 4× faster inference while maintaining 95\%+ accuracy via short-prefix neural recall. 
4) Extensive evaluations on 6 KILT tasks demonstrate state-of-the-art performance, with \method-boosted LLaMA-2 achieving 78.79\% accuracy on FEVER.


\section{\method Framework}

\begin{figure}[!ht]
\centering
\includegraphics[width=1.02\linewidth]{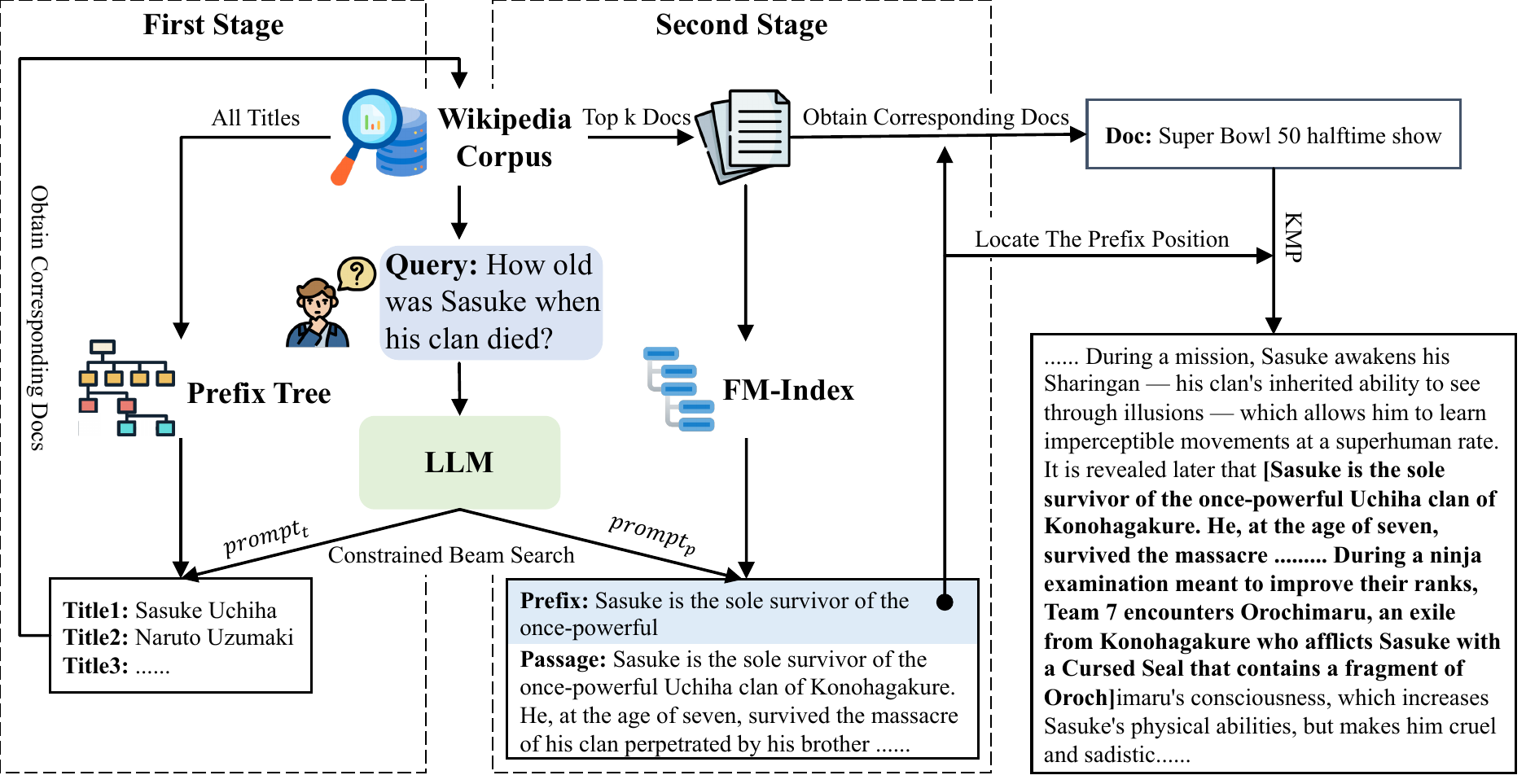} 
\caption{\textbf{\method} Framework. 
First, all Wikipedia titles are stored in a prefix tree, then the LLM is prompted to recall title identifiers under this prefix tree constraint. 
Subsequently, an FM-index is constructed from the top $k$ documents obtained, and the LLM recalls reference passage under the new constraint. 
}
\label{fig2}
\end{figure}

In this section, we detail our two-stage framework, \textbf{\method} (Figure \ref{fig2}). 
In the first stage, we prompt the LLM to recall title identifiers, which serve as candidate documents for the next stage. 
In the second stage, the LLM is prompted to recall reference passage from the documents obtained in the first stage. 
To increase speed, we only recall a short prefix, then locate and extract the reference within the document.
Detailed Prompt can be found in Subsection \ref{sec:prompt}.

\subsection{Stage 1: Coarse-Grained Document Recall}

Recalling fine-grained reference passages directly for knowledge-intensive tasks can be challenging (Subsection \ref{subsec:further}). 
Therefore, we propose a two-stage process. 
First, we retrieve easily-recallable documents (e.g., Wikipedia pages) by leveraging their titles as unique identifiers. 
Using a Trie (Section \ref{trie}) data structure \cite{cormen2022introduction}, we prompt the LLM to recall relevant titles, ensuring they correspond to existing pages.

Given a query $x$ and prompt ${prompt}_{t}(x)$ (e.g., "Question: {}\textbackslash n \textbackslash n The Wikipedia title corresponding to the above question is: \textbackslash n \textbackslash n Title:"), the LLM generates titles, guided by the Trie. 
The Trie, based on previously generated tokens, restricts subsequent token choices to valid prefixes within the set of all Wikipedia titles ($T$), effectively guiding the LLM along valid title paths.

This first stage focuses on efficiently retrieving a set of candidate documents before proceeding to fine-grained passage retrieval within these documents in the second stage.
The score for generating title $t$ given prompt ${prompt}_{t}(x)$ is calculated using the standard implementation from the library:

\begin{equation}
\footnotesize
\begin{aligned}\label{eq1}
&score_{1}(t | {prompt}_{t}(x)) \\&= \frac{\log p_{\theta}(y_{t} | {prompt}_{t}(x))}{|y_{t}|} \\&= \frac{\sum_{i=1}^{l_{t}}\log p_{\theta}(y_{i}|y_{<i},{prompt}_{t}(x))}{l_{t}}
\end{aligned}
\end{equation}
where $y_{t}$ represents the set of tokens in title $t$, $l_{t}$ and $|y_{t}|$ represent the number of tokens used to generate the title, $\theta$ represents the model's parameters.

\subsection{Stage 2: Fine-Grained Passage Recall}
Following the initial identification of relevant Wikipedia pages, we move to fine-grained passage retrieval within those documents. 
We employ the FM-index ((Section \ref{fmindex})) constraint \cite{ferragina2000opportunistic}, a space-efficient data structure enabling fast substring search and supporting retrieval from arbitrary positions.  

After obtaining the top $k$ titles and their documents ($D_k$), we construct a targeted FM-index specifically for $D_k$, reducing the search space. 
These FM-indexes are pre-built for all documents to avoid on-the-fly construction.

We then prompt the LLM with ${prompt}_{p}$ (e.g., "Question: {}\textbackslash n \textbackslash n The answer to the above question can be found in the following Wikipedia paragraph:\textbackslash n \textbackslash n Answer:") to generate a passage $p$.  
The FM-index, based on previously generated tokens, dynamically provides permissible successor tokens, guiding the LLM to generate valid passages from any position within $D_k$. 
We measure the score of the task corresponding passage by using the autoregressive formula to calculate the score:
\begin{equation}
\footnotesize
\begin{aligned}\label{eq2}
&score_{2}(p | {prompt}_{p}(x)) \\&= \frac{\log p_{\theta}(y_{p} | {prompt}_{p}(x))}{|y_{p}|} \\&= \frac{\sum_{i=1}^{l_{p}}\log p_{\theta}(y_{i}|y_{<i},{prompt}_{p}(x))}{l_{p}}
\end{aligned}
\end{equation}
where $y_{p}$ represents the set of tokens in the passage $p$, $\theta$ is the model parameters, $|y_{p}|$ and $l_{p}$ represent the number of tokens generating the passage, usually set between 150 to 200. 
To integrate information generated from both stages, we calculate the weighted sum of the scores from the first and second stages to obtain the final score under input query $x$:
\begin{equation}
\footnotesize
\begin{aligned}\label{eq3}
&score(p|x)=\alpha*score_{1}(t | {prompt}_{t}(x)) \\&+ (1-\alpha)*score_{2}(p | {prompt}_{p}(x))
\end{aligned}
\end{equation}
where $score_{1}(t | {prompt}_{t}(x))$ is the score of the Wikipedia page title $t$ corresponding to the passage $p$, $\alpha$ is a hyperparameter controlling the weight of the two stages. 
Finally, among all recalled passages, the one with the highest $score(p|x)$ value is selected as the best reference.

\subsection{Short Prefix Recall and Localization}

While LLMs excel at recalling long passages, their inference speed hinders practical application. 
To address this, we propose Short Prefix Recall Location (SPRL), aiming to locate passages by recalling only a short prefix.

Initially, given a question $q$, SPRL prompts the LLM to generate a short prefix $p_s$ of length $l_{p_s}$ using the same prompt ($prompt_p$) as in the second stage, significantly reducing generation cost. 
Subsequently, SPRL attempts to identify the document $d$ containing $p_s$ within the document set $D_k$ obtained in the first stage. 
Due to the limited size of $D_k$, $p_s$ typically maps to a unique document $d$.
If $p_s$ is found in multiple documents within $D_k$, the first such document encountered is selected by default, ensuring a deterministic outcome.
Next, using the KMP algorithm, the first starting position $st$ of $p_{s}$ in $d$ is determined, and a complete reference passage $p_{final} = d[st:st + l_{p}]$ is extracted. 
The final score is calculated using Equations \ref{eq2} and \ref{eq3} to select the best reference. 
Experiments (Section \ref{3.2.1}) demonstrate that recalling only the prefix for localization yields effective results.

\section{Experiments}
\label{sec:Experiments}

In this section, we conduct comprehensive experiments on coarse-grained pages, fine-grained passage-level reference evaluation, and downstream tasks to validate the effectiveness of our framework. 
Additionally, we perform further analyses and experiments in the Appendix through Further Analysis, and Case Studies.

\subsection{Experimental Setup}

\textbf{Datasets:} Experiments were conducted on 6 knowledge-sensitive tasks from the KILT benchmark \cite{petroni-etal-2021-kilt} ((Section \ref{sec:datasets})).

\textbf{Evaluation Metrics:} 
R-Precision for page-level retrieval. Answer in Context (percentage of references containing at least one gold answer) for NQ, TriviaQA, and HotpotQA. 
Entity in Context (percentage of references containing at least one gold entity) for other datasets. 
Downstream task metrics followed the official KILT implementations.

\textbf{Baseline Models:} 
We compare with several traditional retrieval models. 
These models all use the passage segmentation from the official KILT as the source for obtaining reference. 
For unsupervised retrieval models, we compare the traditional sparse retrieval model BM25\footnote{We implement BM25 retrieval using the https://github.com/castorini/pyserini repository} \citep{robertson2009probabilistic}, and the dense retrieval model Contriever \citep{izacard2022unsupervised}. 
We also compare with the dense retrieval model DPR\footnote{We conduct experiments with the trained DPR model and preprocessed vector index from the https://github.com/facebookresearch/KILT repository.} \citep{karpukhin-etal-2020-dense} that has been fine-tuned on the full dataset. 
We input the first passage retrieved by the model as the reference context into the LLM, which then reads the relevant reference to answer downstream tasks.

\textbf{Implementation Details:} LLaMA \cite{touvron2023Llama} and LLaMA-2 \cite{touvron2023Llama2} (7b and 13b) were used for reference recall. 
LLaMA-2-13b served as the reading model for downstream tasks. 
We merge the passage fragments from KILT into complete documents, serving as the data source for recall. 
The length of the complete documents is arbitrary. In the recall phase, we always use a beam search generation strategy. 
In the first stage of generation, the beam size is set to 15, and we construct an FM-index containing the top $k=2$ documents. 
In the second stage, the beam size is set to 10, the length of the short prefix is $l_{p_{s}}=16$, and we extract a token length of $l_{p}=150$ as the final reference. 
The weight setting for the two-stage weighted method is $\alpha=0.9$. 
All downstream tasks use greedy decoding. 
The prompts used in the experiments can be found in Appendix \ref{sec:prompt}. 
Experiments were conducted on Tesla A100 40G GPUs.

\begin{table}[t]
\centering
\caption{\label{tab1}
Coarse-grained page-level results (R-Precision). $\star$ denotes full data training. \colorbox{cyan!40}{Cyan} indicates best results, and \colorbox{red!10}{pink} indicates second-best.
}
\resizebox{\linewidth}{!}{
\begingroup
\begin{tabular}{l|cccccc}
\toprule
\multirow{2}{*}{ \textbf{Method} } & \multicolumn{4}{|c}{ \textbf{Open-domain QA} } & \textbf{Fact Check.} & \textbf{Dial.} \\
\cmidrule(lr){2-5} \cmidrule(lr){6-6} \cmidrule(lr){7-7}
& NQ & TriviaQA & HotpotQA & ELI5 & FEVER & WoW \\
\midrule Contriever & 34.72 & 34.28 & 26.14 & 11.02 & 55.64 & 29.67 \\
BM25 & 26.33 & 31.78 & 41.30 & 6.83 & 52.09 & 28.78 \\
DPR$^{\star}$ & 54.74 & 45.68 & 25.46 & \colorbox{cyan!40}{\textbf{16.19}} & 56.61 & 26.62 \\
\midrule \method(LLaMA-7b) & 54.46 & \colorbox{cyan!40}{\textbf{57.03}} & 44.56 & \colorbox{red!10}{\textbf{15.13}} & 76.57 & \colorbox{red!10}{\textbf{52.91}} \\
\method(LLaMA-13b) & 54.42 & 55.53 & \colorbox{red!10}{\textbf{46.30}} & 12.94 & \colorbox{red!10}{\textbf{77.55}} & 34.51 \\
\method(LLaMA-2-7b) & \colorbox{red!10}{\textbf{56.33}} & \colorbox{red!10}{\textbf{56.43}} & 46.20 & 14.60 & 77.29 & 49.61 \\
\method(LLaMA-2-13b) & \colorbox{cyan!40}{\textbf{57.77}} & 54.41 & \colorbox{cyan!40}{\textbf{48.70}} & 15.00 & \colorbox{cyan!40}{\textbf{83.69}} & \colorbox{cyan!40}{\textbf{57.63}} \\
\bottomrule
\end{tabular}
\endgroup
}
\end{table}

\begin{table}[t]
\centering
\caption{\label{tab2}
Fine-grained passage-level results (Answer/Entity in Context for top-1 reference).
}
\resizebox{\linewidth}{!}{
\begingroup
\begin{tabular}{l|cccccc}
\toprule
\multirow{2}{*}{ \textbf{Method} } & \multicolumn{4}{|c}{ \textbf{Open-domain QA} } & \textbf{Fact Check.} & \textbf{Dial.} \\
\cmidrule(lr){2-5} \cmidrule(lr){6-6} \cmidrule(lr){7-7}
& NQ & TriviaQA & HotpotQA & ELI5 & FEVER & WoW \\
\midrule 
& \multicolumn{3}{c}{ Answer in Context } & \multicolumn{3}{c}{ Entity in Context } \\
\cmidrule(lr){1-1} \cmidrule(lr){2-4} \cmidrule(lr){5-7}
Contriever & 19.28 & 37.21 & 11.16 & 12.48 & 40.48 & 45.15 \\
BM25 & 23.65 & 58.87 & \colorbox{red!10}{\textbf{29.45}} & 12.01 & \colorbox{red!10}{\textbf{58.33}} & 50.36 \\
DPR$^{\star}$ & \colorbox{cyan!40}{\textbf{47.94}} & \colorbox{red!10}{\textbf{66.60}} & 20.29 & 14.40 & 41.22 & 45.38 \\
\midrule \method(LLaMA-7b) & 36.87 & 58.48 & 25.55 & \colorbox{red!10}{\textbf{15.99}} & 54.85 & \colorbox{red!10}{\textbf{59.40}} \\
\method(LLaMA-13b) & 37.72 & 60.96 & 26.34 & 14.80 & 55.20 & 50.79 \\
\method(LLaMA-2-7b) & 38.07 & 62.88 & 27.55 & \colorbox{cyan!40}{\textbf{16.85}} & 56.23 & 57.79 \\
\method(LLaMA-2-13b) & \colorbox{red!10}{\textbf{40.82}} & \colorbox{cyan!40}{\textbf{68.20}} & \colorbox{cyan!40}{\textbf{30.04}} & 15.06 & \colorbox{cyan!40}{\textbf{58.42}} & \colorbox{cyan!40}{\textbf{63.43}} \\
\bottomrule
\end{tabular}
\endgroup
}
\end{table}

\begin{table}[t]
\centering
\caption{\label{tab3}
Downstream task results. 
}
\centering
\resizebox{\linewidth}{!}{
\begingroup
\begin{tabular}{l|cccccc}
\toprule
\multirow{2}{*}{ \textbf{Method} } & \multicolumn{4}{|c}{ \textbf{Open-domain QA} } & \textbf{Fact Check.} & \textbf{Dial.} \\
\cmidrule(lr){2-5} \cmidrule(lr){6-6} \cmidrule(lr){7-7}
& NQ & TriviaQA & HotpotQA & ELI5 & FEVER & WoW \\
\midrule 

& \multicolumn{3}{c}{ EM } & \multicolumn{1}{c}{ R-L } & \multicolumn{1}{c}{ ACC } & \multicolumn{1}{c}{ F1 } \\
\cmidrule(lr){1-1} \cmidrule(lr){2-4} \cmidrule(lr){5-5} \cmidrule(lr){6-6} \cmidrule(lr){7-7}

LLaMA-2-13b & 19.74 & 68.71 & 15.64 & 19.46 & 73.23 & 13.90 \\
\midrule Contriever & 24.78 & 69.25 & 20.34 & \colorbox{red!10}{\textbf{20.71}} & 73.61 & 13.96 \\
BM25 & 25.84 & 71.49 & \colorbox{cyan!40}{\textbf{27.23}} & 20.48 & 77.54 & 14.02 \\
DPR$^{\star}$ & \colorbox{cyan!40}{\textbf{33.49}} & \colorbox{red!10}{\textbf{72.68}} & 23.13 & \colorbox{cyan!40}{\textbf{20.75}} & 75.27 & 14.17 \\
\midrule \method(LLaMA-7b) & 29.78 & 70.18 & 24.61 & 20.60 & 78.10 & 14.47 \\
\method(LLaMA-13b) & 29.68 & 71.60 & 25.48 & 20.24 & \colorbox{red!10}{\textbf{78.53}} & 14.33 \\
\method(LLaMA-2-7b) & 29.89 & 70.04 & 25.55 & 20.50 & 78.04 & \colorbox{red!10}{\textbf{14.48}} \\
\method(LLaMA-2-13b) & \colorbox{red!10}{\textbf{31.69}} & \colorbox{cyan!40}{\textbf{72.94}} & \colorbox{red!10}{\textbf{26.13}} & 20.61 & \colorbox{cyan!40}{\textbf{78.79}} & \colorbox{cyan!40}{\textbf{14.77}} \\
\bottomrule
\end{tabular}
\endgroup
}
\end{table}

\subsection{Experimental Results}

\subsubsection{Page-level Results}
\label{3.2.1}

Coarse-grained page-level results, as shown in Table \ref{tab1}, demonstrate that the \method framework, when implemented with Llama-2-13b, achieves the best R-precision scores of 57.77, 48.70, 83.69, and 57.63 on the NQ, HotpotQA, FEVER, and WoW datasets, respectively. 
This significantly surpasses the performance of sparse retrieval BM25 and dense retrieval Contriever in a zero-shot scenario. 
It also shows strong competitive power against the fully trained DPR method, especially on the WoW and FEVER datasets, with improvements of 27.08 and 31.01 points, respectively. 
This result is consistent with the hypothesis that LLMs are powerful in recalling coarse-grained title identifiers, enabling the acquisition of high-quality relevant pages that assist in the subsequent fine-grained recall stage.

\subsubsection{Passage-level Results}
\label{3.2.2}

Fine-grained reference passage results, as shown in Table \ref{tab2}, reveal that the \method framework, when implemented with Llama-2-13b, also achieves the best scores of 68.20, 30.04, 58.42, and 63.43 on the TriviaQA, HotpotQA, FEVER, and WoW datasets, respectively. 
We note that the improvement of the framework in fine-grained reference passage compared to the DPR method is relatively reduced compared to the page-level results. 
This suggests potential for optimization in activating LLMs to recall more detailed and longer reference, presenting a greater challenge compared to recalling shorter title. 
Notably, DPR performs excellently on the NQ dataset, which may relate to its training data format. 
Interestingly, in the HotpotQA dataset, BM25 remains competitive, surpassing dense retrieval methods, possibly due to the longer questions in this dataset leading to more vocabulary overlap. 
\method shows significant progress on the FEVER and WoW datasets, demonstrating the potential and adaptability of LLMs in recalling high-quality reference passage across different task formats. 
Furthermore, the general enhancement in performance with the progression from Llama to Llama-2 and the increase in model size indicates a correlation between the recall ability and the underlying capabilities of LLMs.

\subsubsection{Downstream Task Results}

Downstream task results are presented in Table \ref{tab3}. 
\method, based on Llama-2-13b recalled passage, achieved the best scores of 72.94, 78.79, and 14.77 on the TriviaQA, FEVER, and WoW downstream tasks, respectively, validating the performance of LLM recall references in downstream tasks. 
On the open-domain question answering NQ dataset, although DPR performed excellently after full data training, \method also displayed highly competitive performance. 
On the other hand, in the TriviaQA and HotpotQA datasets, due to the length of the questions, BM25 achieved excellent performance by obtaining more vocabulary overlap, yet \method still achieved comparable or better performance in most cases. 
The unsupervised trained Contriever performed relatively poorly across all tasks, emphasizing the crucial role of supervised training in enhancing the performance of dense retrieval models.


\subsection{Ablation Study}
\label{ablation}

\begin{table}[t]
\centering
\caption{
Ablation study results, with the left half showing the R-Precision at the coarse-grained page level and the right half showing Answer in Context for fine-grained passage.
}
\resizebox{\linewidth}{!}{
\begingroup
\begin{tabular}{l|ccc|ccc}
\toprule
\multirow{2}{*}{ \textbf{Method} } & NQ & TriviaQA & HotpotQA & NQ & TriviaQA & HotpotQA \\
 & \multicolumn{3}{c|}{ R-Precision } & \multicolumn{3}{c}{ Answer in Context } \\
\midrule \method & 57.77 & 54.41 & 48.70 & 40.82 & 68.20 & 30.04 \\
\midrule w/o weight & 51.22 & 49.23 & 48.70 & 39.06 & 66.86 & 28.88 \\
w/o SPRL & 55.30 & 51.50 & 48.70 & 37.43 & 64.64 & 26.18 \\
w/o first stage & 32.22 & 24.87 & 23.36 & 36.27 & 63.33 & 24.16 \\
\bottomrule
\end{tabular}
\endgroup
}
\label{tab_ablation}
\end{table}

In this subsection, we compare methods without weighted scores (w/o weight), without Short Prefix Recall Location (w/o SPRL), and without the first stage of document title recall (w/o first stage).
The results are shown in Table \ref{tab_ablation}.

Without weighted scores, relying solely on the recall scores from the second stage leads to a simultaneous decrease in performance for both coarse and fine-grained results, emphasizing the importance of considering scores from both stages. 
The model, by taking into account title scores, is more capable of selecting the correct document, and within the correct document, it is more likely to choose the correct reference. 
More results on the choice of weighted $\alpha$ can be found in Figure \ref{fig_alpha}.

Without SPRL, recalling longer segments has a minor impact on page-level performance. 
However, it significantly affects the quality of fine-grained reference passage, where longer recall lengths paradoxically lead to decreased performance. 
This result is somewhat counterintuitive and might be due to all document knowledge being stored in the parameters during the pre-training phase, with a short prefix sufficient to locate the required reference. 
Longer references introduce redundancy and noise, thus lowering effectiveness. 
Notably, when using LLaMA-2-13b for recall, recalling complete passages on the NQ dataset takes about 600 minutes, while recalling short prefixes only requires 150 minutes, significantly reducing time costs. 
However, considering that dense retrieval takes about 20 minutes, further optimization of speed remains crucial. 
More experiments on prefix length can be found in Figure \ref{fig_prefix}.

Without the first stage of document title recall, the quality of reference further declines, significantly impacting the quality of page retrieval. 
This indicates that using LLMs to directly recall references across a vast number of documents has considerable limitations and opportunities for improvement. 
The ability of merely prompting LLMs to recall and locate fine-grained reference passage is very limited, making the first stage of recalling document title identifiers crucial.




\section{Related Work}
Traditional retrieval methods rely on sparse (TF-IDF, BM25 \cite{robertson2009probabilistic}) or dense (ORQA \cite{lee2019latent}, DPR \cite{karpukhin-etal-2020-dense}) representations. 
However, dual-encoder dense retrieval faces limitations due to shallow interactions between independently encoded question and passage representations \cite{khattab2020colbert}.

Recent work explores using LLMs to generate identifiers for retrieval, aiming to simplify the process and enhance interaction compared to dual-encoder models. 
These approaches target page titles \cite{cao2021autoregressive}, hierarchical paths \cite{tay2022transformer}, n-grams \cite{bevilacqua2022autoregressive}, multi-hop paths \cite{lee-etal-2022-generative}, multiple identifiers \cite{li-etal-2023-multiview}, or a two-stage approach with passages and URLs \cite{ren-etal-2023-tome,10889045}.
These methods, however, predominantly retrieve predefined text segments, hindering flexible retrieval from arbitrary positions within full documents.

Leveraging LLMs to directly generate relevant knowledge \cite{fang2022leveraging} or augmenting models with LLM-generated context (e.g., GenRead \cite{yu2023generate}, A+B \cite{tang2024a+}) has also shown promise for knowledge-intensive tasks. 
However, hallucination remains a significant challenge \cite{li2023halueval}, potentially providing unreliable or fabricated information.

\section{Conclusion}

This paper introduces \textbf{\method}, a framework utilizing LLMs to independently recall reference passages for knowledge-sensitive tasks. 
Mimicking human information-seeking behavior, the LLM first recalls relevant document pages, and then locates specific passages within them.  
Beam search, constrained by Trie and FM-index structures, ensures that recalled content is a subset of existing texts. 
This framework is adaptable to various open-source LLMs, broadening their potential applications.



\section*{Limitations}

Although \method demonstrates the potential of LLMs to recall reference passage in knowledge-sensitive tasks like humans, its application still faces several limitations. 
Firstly, this framework struggles to surpass the performance of the current SOTA retrieval models, especially those models that have been fine-tuned on specific tasks through supervision. 
In the future, there is a need to explore more effective ways of instruction tuning for recalling under constraints. 
At the same time, \method relies on document title identifiers for phased recall, meaning that its recall capability may be limited for documents lacking clear titles or identifiers.

Moreover, the framework finds it challenging to effectively recall documents that appear less frequently in the pre-training stage. 
This indicates that if a document appears infrequently in the training data of the LLM, or if the document content significantly differs from the training data, \method may encounter difficulties in recalling these documents. 
For the updating of documents and the injection of new knowledge, \method requires additional training to inject this new information into the model parameters.
There is still a need to explore more efficient, lightweight methods for injecting new documents in the future.

\section*{Ethical Considerations}

Our framework ensures that the generated content is entirely derived from reference materials, with Wikipedia as an example in this paper, thus not introducing additional significant ethical issues. 
However, in practical applications, we must ensure that the source document set relied upon is harmless to prevent the spread of inaccurate or harmful information.

\bibliography{custom}

\begin{thebibliography}{31}
\providecommand{\natexlab}[1]{#1}

\bibitem[{Bevilacqua et~al.(2022)Bevilacqua, Ottaviano, Lewis, Yih, Riedel, and Petroni}]{bevilacqua2022autoregressive}
Michele Bevilacqua, Giuseppe Ottaviano, Patrick Lewis, Scott Yih, Sebastian Riedel, and Fabio Petroni. 2022.
\newblock Autoregressive search engines: Generating substrings as document identifiers.
\newblock \emph{Advances in Neural Information Processing Systems}, 35:31668--31683.

\bibitem[{Burrows et~al.(1994)Burrows, L, Taylor, Wheeler, and Wheeler}]{burrows1994block}
Michael Burrows, D~J Wheeler D I G I T~A L, Robert~W. Taylor, David~J. Wheeler, and David Wheeler. 1994.
\newblock A block-sorting lossless data compression algorithm.

\bibitem[{Chiang et~al.(2023)Chiang, Li, Lin, Sheng, Wu, Zhang, Zheng, Zhuang, Zhuang, Gonzalez et~al.}]{vicuna2023}
Wei-Lin Chiang, Zhuohan Li, Zi~Lin, Ying Sheng, Zhanghao Wu, Hao Zhang, Lianmin Zheng, Siyuan Zhuang, Yonghao Zhuang, Joseph~E Gonzalez, et~al. 2023.
\newblock Vicuna: An open-source chatbot impressing gpt-4 with 90\%* chatgpt quality.
\newblock \emph{See https://vicuna. lmsys. org (accessed 14 April 2023)}, 2(3):6.

\bibitem[{Cormen et~al.(2022)Cormen, Leiserson et~al.}]{cormen2022introduction}
Thomas~H Cormen, Charles~E Leiserson, et~al. 2022.
\newblock \emph{Introduction to algorithms}.
\newblock MIT press.

\bibitem[{{De Cao} et~al.(2021){De Cao}, Izacard, Riedel, and Petroni}]{cao2021autoregressive}
Nicola {De Cao}, Gautier Izacard, Sebastian Riedel, and Fabio Petroni. 2021.
\newblock Autoregressive entity retrieval.
\newblock In \emph{9th International Conference on Learning Representations, {ICLR} 2021, Virtual Event, Austria, May 3-7, 2021}.

\bibitem[{Dinan et~al.(2019)Dinan, Roller, Shuster, Fan, Auli, and Weston}]{dinan2018wizard}
Emily Dinan, Stephen Roller, Kurt Shuster, Angela Fan, Michael Auli, and Jason Weston. 2019.
\newblock Wizard of wikipedia: Knowledge-powered conversational agents.
\newblock In \emph{7th International Conference on Learning Representations, {ICLR} 2019, New Orleans, LA, USA, May 6-9, 2019}.

\bibitem[{Fan et~al.(2019)Fan, Jernite, Perez, Grangier, Weston, and Auli}]{fan-etal-2019-eli5}
Angela Fan, Yacine Jernite, Ethan Perez, David Grangier, Jason Weston, and Michael Auli. 2019.
\newblock \href {https://doi.org/10.18653/V1/P19-1346} {{ELI5:} long form question answering}.
\newblock In \emph{Proceedings of the 57th Conference of the Association for Computational Linguistics, {ACL} 2019, Florence, Italy, July 28- August 2, 2019, Volume 1: Long Papers}, pages 3558--3567. Association for Computational Linguistics.

\bibitem[{Fang et~al.(2022)Fang, Wang, Xu, Xu, Sun, Zhu, and Zeng}]{fang2022leveraging}
Yuwei Fang, Shuohang Wang, Yichong Xu, Ruochen Xu, Siqi Sun, Chenguang Zhu, and Michael Zeng. 2022.
\newblock Leveraging knowledge in multilingual commonsense reasoning.
\newblock In \emph{Findings of the Association for Computational Linguistics: {ACL} 2022, Dublin, Ireland, May 22-27, 2022}, pages 3237--3246. Association for Computational Linguistics.

\bibitem[{Ferragina and Manzini(2000)}]{ferragina2000opportunistic}
Paolo Ferragina and Giovanni Manzini. 2000.
\newblock Opportunistic data structures with applications.
\newblock In \emph{41st Annual Symposium on Foundations of Computer Science, {FOCS} 2000, 12-14 November 2000, Redondo Beach, California, {USA}}, pages 390--398. {IEEE} Computer Society.

\bibitem[{Izacard et~al.(2022)Izacard, Caron, Hosseini, Riedel, Bojanowski, Joulin, and Grave}]{izacard2022unsupervised}
Gautier Izacard, Mathilde Caron, Lucas Hosseini, Sebastian Riedel, Piotr Bojanowski, Armand Joulin, and Edouard Grave. 2022.
\newblock Unsupervised dense information retrieval with contrastive learning.
\newblock \emph{Transactions on Machine Learning Research}, 2022.

\bibitem[{Izacard and Grave(2021)}]{izacard-grave-2021-leveraging}
Gautier Izacard and Edouard Grave. 2021.
\newblock Leveraging passage retrieval with generative models for open domain question answering.
\newblock In \emph{Proceedings of the 16th Conference of the European Chapter of the Association for Computational Linguistics: Main Volume, {EACL} 2021, Online, April 19 - 23, 2021}, pages 874--880. Association for Computational Linguistics.

\bibitem[{Joshi et~al.(2017)Joshi, Choi, Weld, and Zettlemoyer}]{joshi-etal-2017-triviaqa}
Mandar Joshi, Eunsol Choi, Daniel~S. Weld, and Luke Zettlemoyer. 2017.
\newblock Trivia{QA}: {A} large scale distantly supervised challenge dataset for reading comprehension.
\newblock In \emph{Proceedings of the 55th Annual Meeting of the Association for Computational Linguistics, {ACL} 2017, Vancouver, Canada, July 30 - August 4, Volume 1: Long Papers}, pages 1601--1611. Association for Computational Linguistics.

\bibitem[{Karpukhin et~al.(2020)Karpukhin, Oguz, Min, Lewis, Wu, Edunov, Chen, and Yih}]{karpukhin-etal-2020-dense}
Vladimir Karpukhin, Barlas Oguz, Sewon Min, Patrick S.~H. Lewis, Ledell Wu, Sergey Edunov, Danqi Chen, and Wen{-}tau Yih. 2020.
\newblock Dense passage retrieval for open-domain question answering.
\newblock In \emph{Proceedings of the 2020 Conference on Empirical Methods in Natural Language Processing, {EMNLP} 2020, Online, November 16-20, 2020}, pages 6769--6781. Association for Computational Linguistics.

\bibitem[{Khattab et~al.(2021)Khattab, Potts, and Zaharia}]{khattab-etal-2021-relevance}
Omar Khattab, Christopher Potts, and Matei Zaharia. 2021.
\newblock Relevance-guided supervision for {O}pen{QA} with {C}ol{BERT}.
\newblock \emph{Transactions of the Association for Computational Linguistics}, 9:929--944.

\bibitem[{Khattab and Zaharia(2010)}]{khattab2020colbert}
Omar Khattab and Matei Zaharia. 2010.
\newblock Colbert: Efficient and effective passage search via contextualized late interaction over {BERT}.
\newblock In \emph{Proceedings of the 43rd International {ACM} {SIGIR} conference on research and development in Information Retrieval, {SIGIR} 2020, Virtual Event, China, July 25-30, 2020}, pages 39--48. {ACM}.

\bibitem[{Kwiatkowski et~al.(2019)Kwiatkowski, Palomaki, Redfield, Collins, Parikh, Alberti, Epstein, Polosukhin, Devlin, Lee, Toutanova, Jones, Kelcey, Chang, Dai, Uszkoreit, Le, and Petrov}]{kwiatkowski-etal-2019-natural}
Tom Kwiatkowski, Jennimaria Palomaki, Olivia Redfield, Michael Collins, Ankur~P. Parikh, Chris Alberti, Danielle Epstein, Illia Polosukhin, Jacob Devlin, Kenton Lee, Kristina Toutanova, Llion Jones, Matthew Kelcey, Ming{-}Wei Chang, Andrew~M. Dai, Jakob Uszkoreit, Quoc Le, and Slav Petrov. 2019.
\newblock Natural questions: a benchmark for question answering research.
\newblock \emph{Transactions of the Association for Computational Linguistics}, 7:452--466.

\bibitem[{Lee et~al.(2022)Lee, Yang, Oh, and Seo}]{lee-etal-2022-generative}
Hyunji Lee, Sohee Yang, Hanseok Oh, and Minjoon Seo. 2022.
\newblock Generative multi-hop retrieval.
\newblock In \emph{Proceedings of the 2022 Conference on Empirical Methods in Natural Language Processing, {EMNLP} 2022, Abu Dhabi, United Arab Emirates, December 7-11, 2022}, pages 1417--1436. Association for Computational Linguistics.

\bibitem[{Lee et~al.(2019)Lee, Chang, and Toutanova}]{lee2019latent}
Kenton Lee, Ming{-}Wei Chang, and Kristina Toutanova. 2019.
\newblock Latent retrieval for weakly supervised open domain question answering.
\newblock In \emph{Proceedings of the 57th Conference of the Association for Computational Linguistics, {ACL} 2019, Florence, Italy, July 28- August 2, 2019, Volume 1: Long Papers}, pages 6086--6096. Association for Computational Linguistics.

\bibitem[{Li et~al.(2023{\natexlab{a}})Li, Cheng, Zhao, Nie, and Wen}]{li2023halueval}
Junyi Li, Xiaoxue Cheng, Xin Zhao, Jian{-}Yun Nie, and Ji{-}Rong Wen. 2023{\natexlab{a}}.
\newblock Halueval: {A} large-scale hallucination evaluation benchmark for large language models.
\newblock In \emph{Proceedings of the 2023 Conference on Empirical Methods in Natural Language Processing, {EMNLP} 2023, Singapore, December 6-10, 2023}, pages 6449--6464. Association for Computational Linguistics.

\bibitem[{Li et~al.(2023{\natexlab{b}})Li, Yang, Wang, Wei, and Li}]{li-etal-2023-multiview}
Yongqi Li, Nan Yang, Liang Wang, Furu Wei, and Wenjie Li. 2023{\natexlab{b}}.
\newblock Multiview identifiers enhanced generative retrieval.
\newblock In \emph{Proceedings of the 61st Annual Meeting of the Association for Computational Linguistics (Volume 1: Long Papers), {ACL} 2023, Toronto, Canada, July 9-14, 2023}, pages 6636--6648. Association for Computational Linguistics.

\bibitem[{Petroni et~al.(2021)Petroni, Piktus, Fan, Lewis, Yazdani, De~Cao, Thorne, Jernite, Karpukhin, Maillard, Plachouras, Rockt{\"a}schel, and Riedel}]{petroni-etal-2021-kilt}
Fabio Petroni, Aleksandra Piktus, Angela Fan, Patrick Lewis, Majid Yazdani, Nicola De~Cao, James Thorne, Yacine Jernite, Vladimir Karpukhin, Jean Maillard, Vassilis Plachouras, Tim Rockt{\"a}schel, and Sebastian Riedel. 2021.
\newblock {KILT}: a benchmark for knowledge intensive language tasks.
\newblock In \emph{Proceedings of the 2021 Conference of the North American Chapter of the Association for Computational Linguistics: Human Language Technologies}.

\bibitem[{Ren et~al.(2023)Ren, Zhao, Liu, Wu, Wen, and Wang}]{ren-etal-2023-tome}
Ruiyang Ren, Wayne~Xin Zhao, Jing Liu, Hua Wu, Ji{-}Rong Wen, and Haifeng Wang. 2023.
\newblock {TOME:} {A} two-stage approach for model-based retrieval.
\newblock In \emph{Proceedings of the 61st Annual Meeting of the Association for Computational Linguistics (Volume 1: Long Papers), {ACL} 2023, Toronto, Canada, July 9-14, 2023}, pages 6102--6114. Association for Computational Linguistics.

\bibitem[{Robertson and Zaragoza(2009)}]{robertson2009probabilistic}
Stephen~E. Robertson and Hugo Zaragoza. 2009.
\newblock The probabilistic relevance framework: {BM25} and beyond.
\newblock \emph{Foundations and Trends{\textregistered} in Information Retrieval}, 3(4):333--389.

\bibitem[{Tang et~al.(2024)Tang, Cao, Ying, Wang, Zhao, Liao, and Zhou}]{tang2024a+}
Wei Tang, Yixin Cao, Jiahao Ying, Bo~Wang, Yuyue Zhao, Yong Liao, and Pengyuan Zhou. 2024.
\newblock A+ b: A general generator-reader framework for optimizing llms to unleash synergy potential.
\newblock \emph{arXiv preprint arXiv:2406.03963}.

\bibitem[{Tay et~al.(2022)Tay, Tran, Dehghani, Ni, Bahri, Mehta, Qin, Hui, Zhao, Gupta, Schuster, Cohen, and Metzler}]{tay2022transformer}
Yi~Tay, Vinh Tran, Mostafa Dehghani, Jianmo Ni, Dara Bahri, Harsh Mehta, Zhen Qin, Kai Hui, Zhe Zhao, Jai~Prakash Gupta, Tal Schuster, William~W. Cohen, and Donald Metzler. 2022.
\newblock Transformer memory as a differentiable search index.
\newblock In \emph{Advances in Neural Information Processing Systems 35: Annual Conference on Neural Information Processing Systems 2022, NeurIPS 2022, New Orleans, LA, USA, November 28 - December 9, 2022}.

\bibitem[{Thorne et~al.(2018)Thorne, Vlachos, Christodoulopoulos, and Mittal}]{thorne-etal-2018-fever}
James Thorne, Andreas Vlachos, Christos Christodoulopoulos, and Arpit Mittal. 2018.
\newblock {FEVER:} a large-scale dataset for fact extraction and {VER}ification.
\newblock In \emph{Proceedings of the 2018 Conference of the North American Chapter of the Association for Computational Linguistics: Human Language Technologies, {NAACL-HLT} 2018, New Orleans, Louisiana, USA, June 1-6, 2018, Volume 1 (Long Papers)}, pages 809--819. Association for Computational Linguistics.

\bibitem[{Touvron et~al.(2023{\natexlab{a}})Touvron, Lavril, Izacard, Martinet, Lachaux, Lacroix, Rozi{\`{e}}re, Goyal, Hambro, Azhar, Rodriguez, Joulin, Grave, and Lample}]{touvron2023Llama}
Hugo Touvron, Thibaut Lavril, Gautier Izacard, Xavier Martinet, Marie{-}Anne Lachaux, Timoth{\'{e}}e Lacroix, Baptiste Rozi{\`{e}}re, Naman Goyal, Eric Hambro, Faisal Azhar, Aur{\'{e}}lien Rodriguez, Armand Joulin, Edouard Grave, and Guillaume Lample. 2023{\natexlab{a}}.
\newblock {LLaMA}: Open and efficient foundation language models.
\newblock \emph{CoRR}, abs/2302.13971.

\bibitem[{Touvron et~al.(2023{\natexlab{b}})Touvron, Martin, Stone, Albert, Almahairi, Babaei, Bashlykov, Batra, Bhargava, Bhosale, Bikel, Blecher, Canton{-}Ferrer, Chen, Cucurull et~al.}]{touvron2023Llama2}
Hugo Touvron, Louis Martin, Kevin Stone, Peter Albert, Amjad Almahairi, Yasmine Babaei, Nikolay Bashlykov, Soumya Batra, Prajjwal Bhargava, Shruti Bhosale, Dan Bikel, Lukas Blecher, Cristian Canton{-}Ferrer, Moya Chen, Guillem Cucurull, et~al. 2023{\natexlab{b}}.
\newblock {LLaMA} 2: Open foundation and fine-tuned chat models.
\newblock \emph{CoRR}, abs/2307.09288.

\bibitem[{Yang et~al.(2018)Yang, Qi, Zhang, Bengio, Cohen, Salakhutdinov, and Manning}]{yang-etal-2018-hotpotqa}
Zhilin Yang, Peng Qi, Saizheng Zhang, Yoshua Bengio, William~W. Cohen, Ruslan Salakhutdinov, and Christopher~D. Manning. 2018.
\newblock Hotpotqa: {A} dataset for diverse, explainable multi-hop question answering.
\newblock In \emph{Proceedings of the 2018 Conference on Empirical Methods in Natural Language Processing, Brussels, Belgium, October 31 - November 4, 2018}, pages 2369--2380. Association for Computational Linguistics.

\bibitem[{Yu et~al.(2023)Yu, Iter, Wang, Xu, Ju, Sanyal, Zhu, Zeng, and Jiang}]{yu2023generate}
Wenhao Yu, Dan Iter, Shuohang Wang, Yichong Xu, Mingxuan Ju, Soumya Sanyal, Chenguang Zhu, Michael Zeng, and Meng Jiang. 2023.
\newblock Generate rather than retrieve: Large language models are strong context generators.
\newblock In \emph{The Eleventh International Conference on Learning Representations, {ICLR} 2023, Kigali, Rwanda, May 1-5, 2023}. OpenReview.net.

\bibitem[{Yue et~al.(2025)Yue, Xu, Ma, Du, Ding, Han, Zhang, and Zhang}]{10889045}
Chongjian Yue, Xinrun Xu, Xiaojun Ma, Lun Du, Zhiming Ding, Shi Han, Dongmei Zhang, and Qi~Zhang. 2025.
\newblock \href {https://doi.org/10.1109/ICASSP49660.2025.10889045} {Extract information from hybrid long documents leveraging llms: A framework and dataset}.
\newblock In \emph{ICASSP 2025 - 2025 IEEE International Conference on Acoustics, Speech and Signal Processing (ICASSP)}, pages 1--5.

\end{thebibliography}

\clearpage

\appendix

\section{Preliminary}
\label{sec:preliminary}

\subsection{Trie}
\label{trie}

\begin{figure}[!ht]
    \centering
    \begin{subfigure}{0.45\linewidth}
        \centering
        \includegraphics[width=\linewidth]{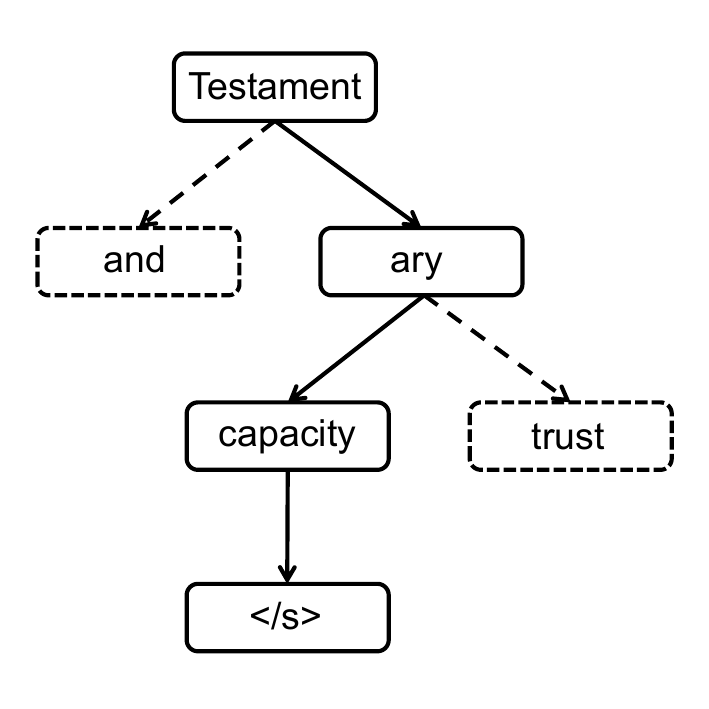}
        \caption{Title Generation with Prefix Tree.}
        \label{fig_index:sub1}
    \end{subfigure}
    \hfill
    \begin{subfigure}{0.45\linewidth}
        \centering
        \includegraphics[width=\linewidth]{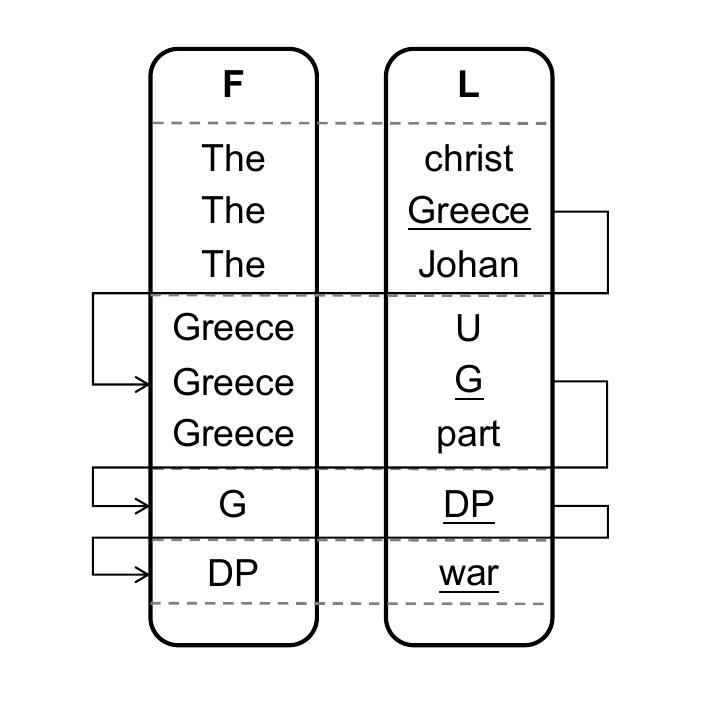}
        \caption{Passage Prefix Generation with FM-index.}
        \label{fig_index:sub2}
    \end{subfigure}
    \caption{Constrained Decoding Methods: (a) Shows the process of an LLM generating title identifiers using a prefix tree. (b) Shows the process of an LLM generating passage prefixes in a document set via FM-index.}
    \label{fig:main}
\end{figure}

The Trie \citep{cormen2022introduction}, also known as a dictionary tree or prefix tree, is a tree-like data structure used to store an associative array where the keys are usually strings. 
Unlike a binary search tree, keys in a Trie are not stored directly within the nodes; instead, they are determined by the node's position in the tree. All descendants of a node have the same prefix, associated with the string corresponding to that node.

The overall process during constrained decoding using a Trie is shown in Figure \ref{fig_index:sub1}. 
Taking the generation of the title "Testamentary Capacity" as an example, the LLM first selects "Testament" from the set of token strings that start all titles. 
Subsequently, we can obtain the set of token strings \{and, ary\} following the string "Testament". 
After the LLM selects "ary", we get the prefix string "Testamentary", and finally continue to select new strings from the next set of token strings until the end-of-sequence token \textless/s\textgreater is encountered, ceasing generation.

\subsection{FM-Index}
\label{fmindex}

The FM-index \citep{ferragina2000opportunistic} is a data structure used for text retrieval that can store text efficiently with linear space complexity and support fast substring search operations. 
It is constructed based on the Burrows-Wheeler Transform (BWT) \citep{burrows1994block}. 
BWT is a method that converts a string into a form that is easy to compress. 
Given a string, BWT produces a transformed string through the following steps: generate all cyclic shifts of the string, sort all these shifts lexicographically, take the last character of each sorted shifted string to form a new string, which is the BWT result. 
For example, for the string "CABAC", the process of building the FM-index is as follows:

\begin{center}
$\begin{array}{cccccc}
\mathbf{F} & & & & & \mathbf{L} \\
\$^6 & C & A & B & A & C^5 \\
A^2 & B & A & C & \$ & C^1 \\
A^4 & C & \$ & C & A & B^3 \\
B^3 & A & C & \$ & C & A^2 \\
C^5 & \$ & C & A & B & A^4 \\
C^1 & A & B & A & C & \$^6
\end{array}$
\end{center}

where \$ is a special string termination token, the numbers in the upper right corner of the letters in the F and L columns are the corresponding position index numbers. 
The FM-index explicitly stores two main parts: the F column and the L column. 
The F column is the lexicographically sorted characters of the transformed string, and the L column is the result of BWT. 
In addition, it stores additional position information to recover the original string from the BWT result. 
When we want to query a substring, the FM-index starts from the last character of the substring, using the information in the F column and the L column to gradually narrow down the possible position range until the exact position of the substring is determined or the substring is determined to be non-existent.

The overall process during constrained decoding using FM-index is shown in Figure \ref{fig_index:sub2}. 
Considering the generated prefix "The Greece GDP warrants are not technically bonds as investors do" for example, it first starts from the string "The" generated from all corpus, and gets its corresponding L column string set \{christ, Greece, Johan\}. 
After "Greece" is selected by the LLM, we can get the next set \{U, G, part\}, and continue the iteration until reaching the set maximum prefix length to stop generating.

\section{Additional Details for Experiments}

\subsection{Datasets}
\label{sec:datasets}

\begin{table*}[!ht]
\centering
\begin{tabular}{l|llll}
\toprule
\textbf{Dataset} & \textbf{Task} & \textbf{Input Format} & \textbf{Output Format} & \textbf{Size} \\
\midrule
NQ \citep{kwiatkowski-etal-2019-natural} & Open-domain QA & Question & Extractive & 2837 \\
HotpotQA \citep{yang-etal-2018-hotpotqa} & Open-domain QA & Question & Short Abstractive & 5600 \\
TriviaQA \citep{joshi-etal-2017-triviaqa} & Open-domain QA& Question & Extractive & 5359 \\
ELI5 \citep{fan-etal-2019-eli5} & Open-domain QA & Question & Long Abstractive & 1507 \\
FEVER \citep{thorne-etal-2018-fever}& Fact Checking & Claim & Classification & 10444 \\
WoW \citep{dinan2018wizard}& Dialogue & Conversation & Long Abstractive & 3054 \\
\bottomrule
\end{tabular}
\caption{
Additional details of the datasets. 
}
\label{tab_datasets}
\end{table*}

We conduct extensive experiments on 6 knowledge-sensitive tasks from the KILT benchmark \citep{petroni-etal-2021-kilt}. 
These tasks include open-domain QA tasks such as NQ \citep{kwiatkowski-etal-2019-natural}, TriviaQA \citep{joshi-etal-2017-triviaqa}, HotpotQA \citep{yang-etal-2018-hotpotqa}, and ELI5 \citep{fan-etal-2019-eli5}, the fact-checking task FEVER \citep{thorne-etal-2018-fever}, and the open-domain dialogue system WoW \citep{dinan2018wizard}.
All experiments are tested using the public validation set as divided in the official KILT. 
Additional details of the datasets are presented in Table \ref{tab_datasets}. 
All the data used in this paper come from the KILT benchmark\citep{petroni-etal-2021-kilt}, and KILT is MIT licensed\footnote{https://opensource.org/licenses/MIT}.
We evaluate the quality of coarse-grained pages and fine-grained reference passage, as well as the enhancement of reference for downstream tasks.

\subsection{Prompt}
\label{sec:prompt}

In this subsection, we introduce the prompt used in the first stage for recalling coarse-grained title identifiers, in the second stage for recalling fine-grained reference passage, and in downstream tasks.

\subsubsection{Prompt for the First Stage}

\begin{itemize}
    \item Open-domain QA: "Question: \{\}\textbackslash n \textbackslash n  The Wikipedia article corresponding to the above question is:\textbackslash n \textbackslash n  Title:"
    
    \item Fact Verification: "Claim: \{\}\textbackslash n \textbackslash n The Wikipedia article corresponding to the above claim is:\textbackslash n \textbackslash n Title:"
    
    \item Open-domain Dialogue System: "Conversation: \{\}\textbackslash n \textbackslash n The Wikipedia article corresponding to the above conversation is:\textbackslash n \textbackslash n Title:"
    
\end{itemize}

\subsubsection{Prompt for the Second Stage}

\begin{itemize}

    \item Open-domain QA: "Question: \{\}\textbackslash n \textbackslash n  The Wikipedia paragraph to answer the above question is:\textbackslash n \textbackslash n  Answer:"

    \item Fact Verification: "Claim: \{\}\textbackslash n \textbackslash n The Wikipedia paragraph to support or refute the above claim is:\textbackslash n \textbackslash n Answer:"

    \item Open-domain Dialogue System: "Conversation: \{\}\textbackslash n \textbackslash n The Wikipedia paragraph to answer the above conversation is:\textbackslash n \textbackslash n Answer:"
\end{itemize}

\subsubsection{Prompt for Reading Comprehension}

\begin{itemize}

    \item Open-domain QA (NQ, TriviaQA, HotpotQA): "Refer to the passage below and answer the following question with just a few words.\textbackslash n Passage: \{\}\textbackslash n Q: \{\}\textbackslash n A: The answer is"

    \item Open-domain QA (ELI5): "Refer to the passage below and answer the following question in detail.\textbackslash n Passage: \{\}\textbackslash n Q: \{\}\textbackslash n A:"

    \item Fact Verification: "background: \{\}\textbackslash n claim: \{\}\textbackslash n Q: Is the claim true or false?\textbackslash n A:"

    \item Open-domain Dialogue System: "background: \{\}\textbackslash n \{\}\textbackslash n "

\end{itemize}



\begin{table*}[t]
\begin{tabular}{l|ccc|ccc}
\toprule
\textbf{Method} & \textbf{NQ} & \textbf{TriviaQA} & \textbf{HotpotQA} & \textbf{NQ} & \textbf{TriviaQA} & \textbf{HotpotQA} \\
\midrule & \multicolumn{3}{c|}{ R-Precision } & \multicolumn{3}{c}{ Answer in Context } \\
\midrule \method(LLaMA-7b) & 54.46 & 57.03 & 44.56 & 36.87 & 58.48 & 25.55 \\
\method(Vicuna-1.3-7b) & 48.47 & 47.99 & 40.79 & 35.28 & 56.41 & 23.75 \\
\midrule \method(LLaMA-13b) & 54.42 & 55.53 & 46.30 & 37.72 & 60.96 & 26.34 \\
\method(Vicuna-1.3-13b) & 52.73 & 46.61 & 43.41 & 36.55 & 67.29 & 26.63 \\
\midrule \method(LLaMA-2-7b) & 56.33 & 56.43 & 46.20 & 38.07 & 62.88 & 27.55 \\
\method(LLaMA-2-chat-7b) & 3.31 & 1.12 & 0.98 & 4.09 & 3.97 & 3.43 \\
\method(Vicuna-1.5-7b) & 50.76 & 51.73 & 41.23 & 34.16 & 55.98 & 24.43 \\
\midrule \method(LLaMA-2-13b) & 57.77 & 54.41 & 48.70 & 40.82 & 68.20 & 30.04 \\
\method(LLaMA-2-chat-13b) & 1.94 & 1.60 & 1.55 & 6.38 & 7.93 & 4.71 \\
\method(Vicuna-1.5-13b) & 52.24 & 56.34 & 45.90 & 37.22 & 63.24 & 27.14 \\
\bottomrule
\end{tabular}
\caption{
On the NQ, TriviaQA, and HotpotQA datasets, experimental results after general fine-tuning of the model are presented. 
The left side shows the page-level R-Precision, while the right side displays the passage-level Answer in Context.
}
\label{tab_finetune}
\end{table*}

\begin{figure*}[!ht]
\centering
\includegraphics[width=0.65\linewidth]{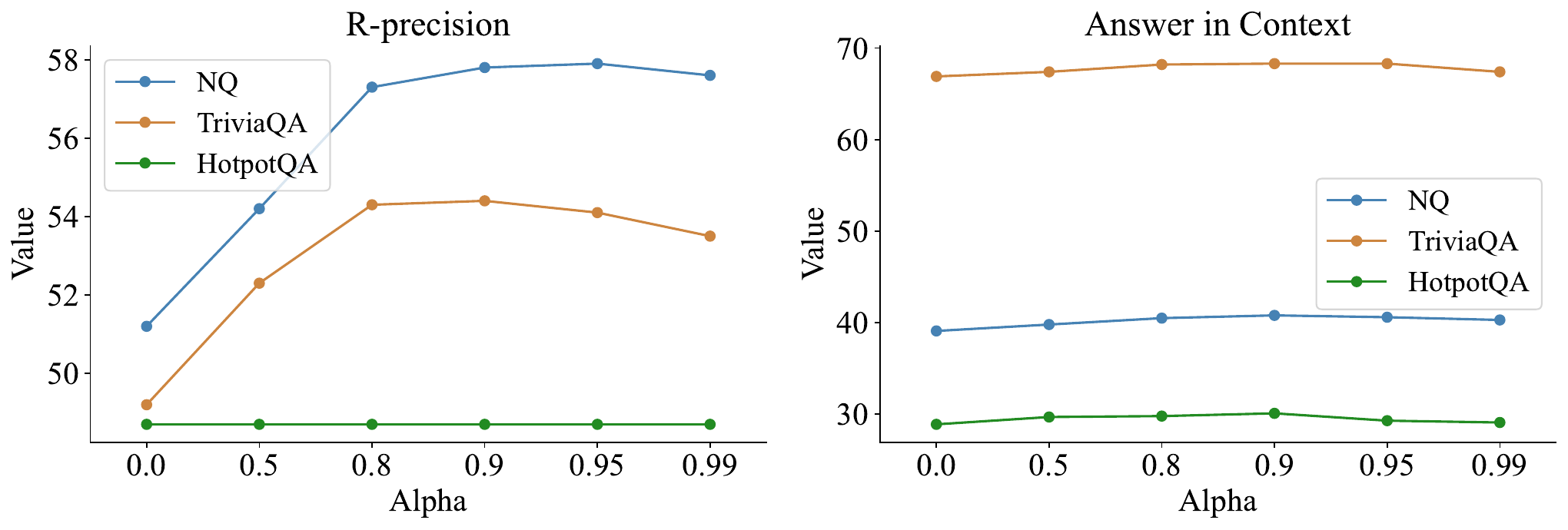} 
\caption{On the NQ, TriviaQA, and HotpotQA datasets, the page-level and passage-level experimental results for LLaMA-2-13b when setting $\alpha$ to \{0.0,0.5,0.8,0.9,0.95,0.99\}.}
\label{fig_alpha}
\end{figure*}

\begin{figure*}[!ht]
\centering
\includegraphics[width=0.65\linewidth]{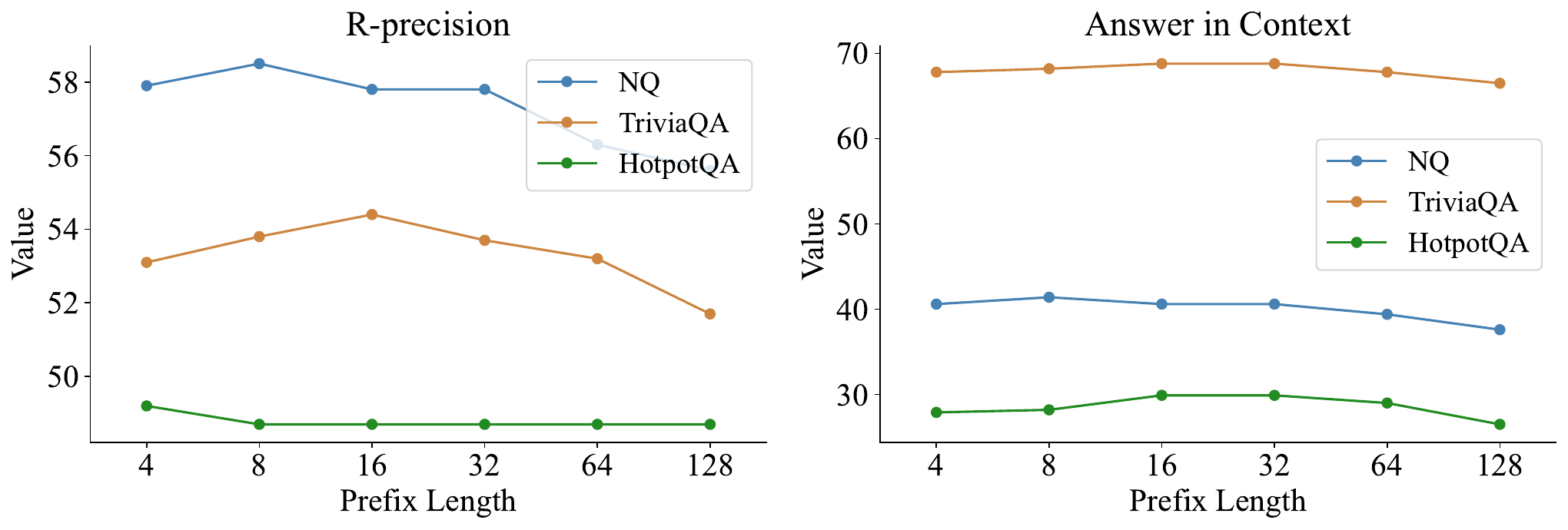} 
\caption{On the NQ, TriviaQA, and HotpotQA datasets, the page-level and passage-level experimental results for LLaMA-2-13b with different prefix token lengths $l_{p_{s}}$ set to \{4,8,16,32,64,128\}.}
\label{fig_prefix}
\end{figure*}

\begin{figure*}[!ht]
\centering
\includegraphics[width=0.65\linewidth]{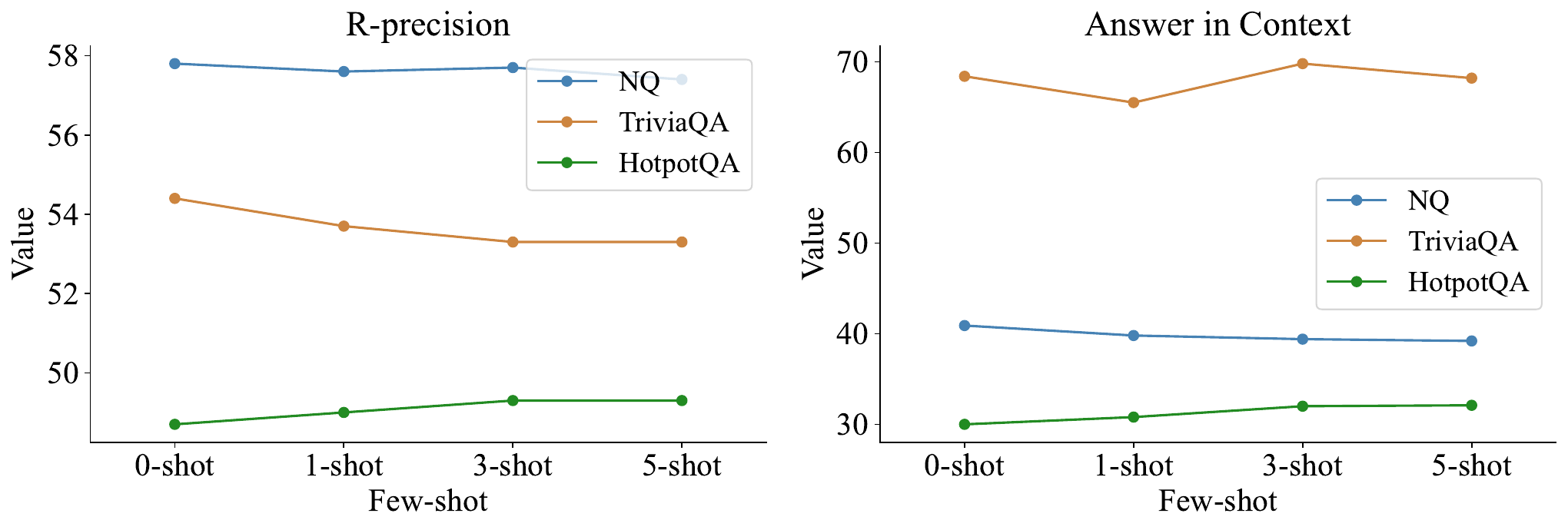}
\caption{On the NQ, TriviaQA, and HotpotQA datasets, the page-level and passage-level experimental results for LLaMA-2-13b under \{0,1,3,5\}-shot few-shot prompt.}
\label{fig_few-shot}
\end{figure*}

\begin{figure*}[!ht]
\centering
\includegraphics[width=0.65\linewidth]{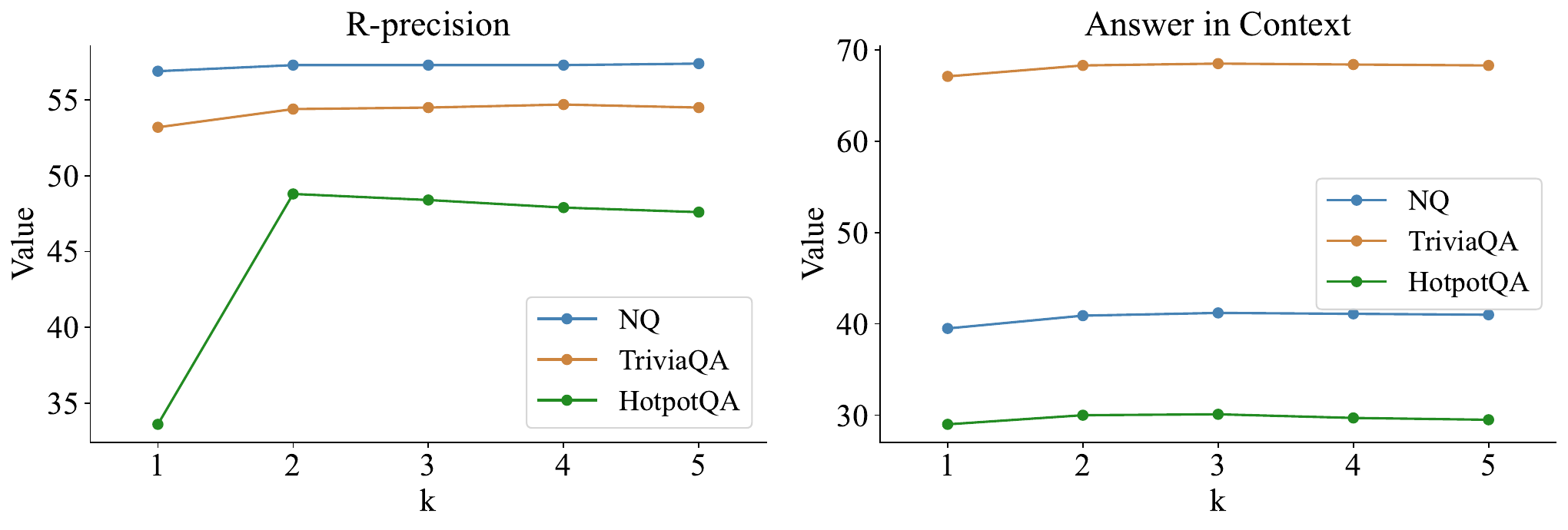} 
\caption{On the NQ, TriviaQA, and HotpotQA datasets, the page-level and passage-level experimental results for LLaMA-2-13b with the number of documents selected in the first stage $k$ set to \{1,2,3,4,5\}.}
\label{fig_topk}
\end{figure*}

\begin{figure*}[!ht]
\centering
\includegraphics[width=0.65\linewidth]{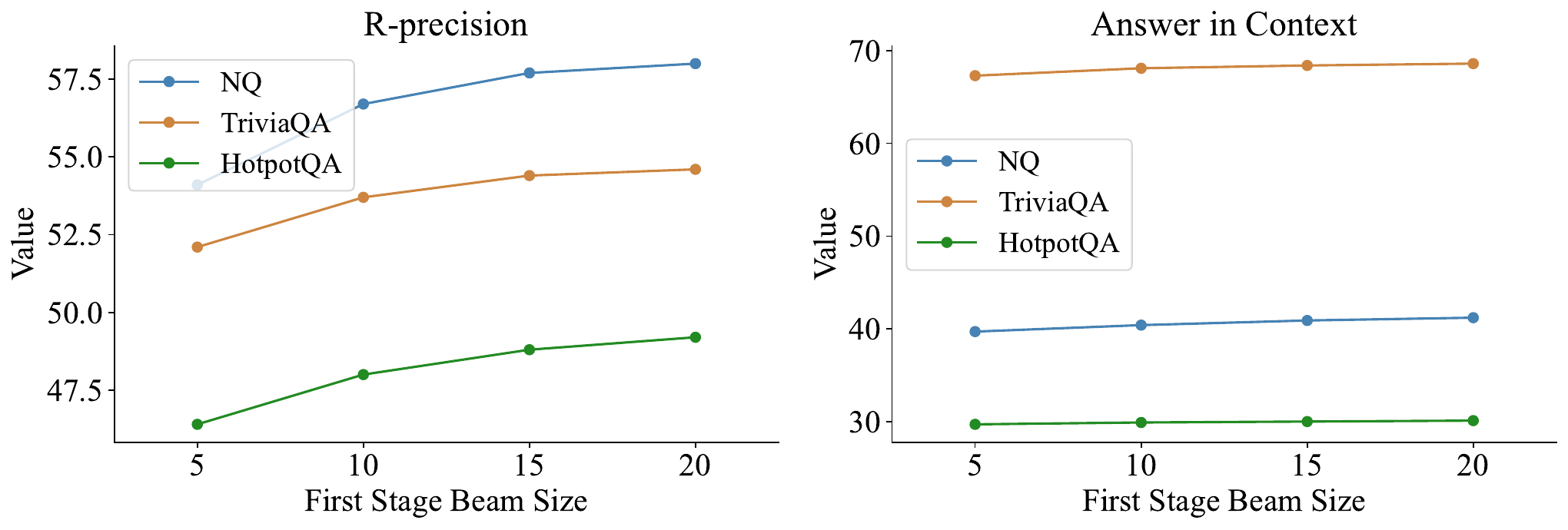} 
\caption{On the NQ, TriviaQA, and HotpotQA datasets, the page-level and passage-level experimental results for LLaMA-2-13b with different beam search sizes \{4,8,16,32,64,128\} set for the first stage.}
\label{fig_first_beam_size}
\end{figure*}

\begin{figure*}[!ht]
\centering
\includegraphics[width=0.65\linewidth]{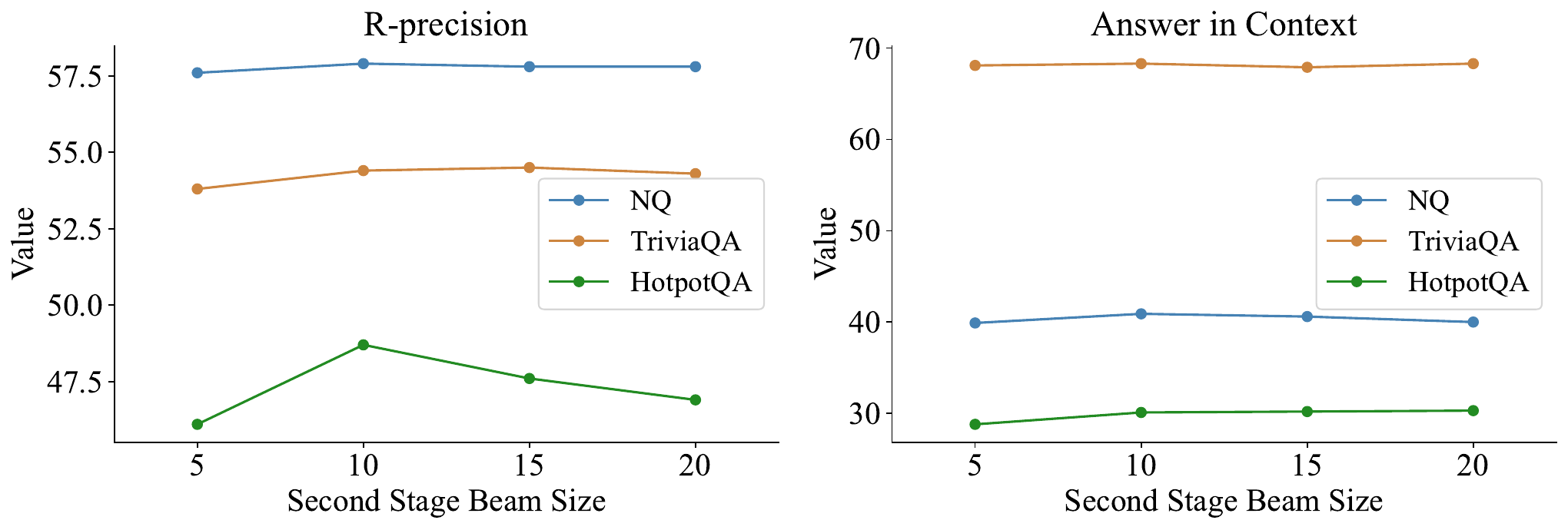}
\caption{On the NQ, TriviaQA, and HotpotQA datasets, the page-level and passage-level experimental results for LLaMA-2-13b with different beam search sizes \{4,8,16,32,64,128\} set for the second stage.}
\label{fig_second_beam_size}
\end{figure*}

\subsection{Further Analysis}
\label{subsec:further}

\textbf{Different Values of Alpha} In Figure \ref{fig_alpha}, we compare the experimental results of \method when implemented based on Llama-2-13b with different $\alpha$ values. 
When $\alpha=0.0$, it's equivalent to not having a two-stage weighted method, relying only on the scores from the second stage's fine-grained passage recall, resulting in the selection of suboptimal reference. 
With the increase of $\alpha$, the model sees improvements in both page-level and passage-level results, proving the importance of the first stage document scores for the final reference selection. 
However, when $\alpha$ reaches 0.95 and continues to increase, the final performance actually decreases to some extent, indicating the need to find a balance between the two for better results.

\textbf{Different Prefix Lengths} In Figure \ref{fig_prefix}, we conduct experiments with \method recalling different numbers of prefix tokens based on Llama-2-13b. 
We observe that longer prefix lengths do not bring additional performance improvements; on the contrary, they lead to a decrease in performance. 
Existing LLMs still perform better when generating shorter passages under constraints; longer passages introduce additional noise, resulting in decreased performance. 
However, overly short prefixes might also lack sufficient information, leading to an inability to accurately select the desired passage as a reference.

\textbf{After General Fine-tuning of LLMs.} We also test the Vicuna model \citep{vicuna2023} and the LLaMA-2-chat model refined through reinforcement learning from human feedback \citep{touvron2023Llama2}, both of which underwent general fine-tuning. This general fine-tuning did not significantly enhance the performance of LLMs in recalling and locating reference. 
This may be due to the paradigm difference between the fine-tuning data and the recall location task, coupled with the fact that most knowledge was already acquired during the pre-training phase. 
By creating more diverse recall instruction tuning data, further improvements in model performance might be achieved. 
Detailed results can be found in Table \ref{tab_finetune}.

\textbf{Impact of Few-Shot.} We explore adding few-shot prompt in the second stage of fine-grained recall and observed its impact on overall performance. This approach brought slight improvements only in the HotpotQA dataset, while showing a slight decline in NQ and TriviaQA. 
Importantly, adding more few-shot examples significantly reduced generation speed. 
This suggests that, although few-shot prompting offers a potential path for improvement, extensive exploration is still needed to devise more effective prompting methods. 
Detailed results can be found in Figure \ref{fig_few-shot}.

\textbf{Impact of Document Selection ($k$) of First Stage Documents.}
In Figure \ref{fig_topk}, we conduct experiments to compare the effect of selecting different numbers of first-stage documents (denoted as $k$) in \method when implemented based on LLaMA-2-13b. 
We observe that the impact of $k$ on the final performance is not significant, as the necessary effective reference passages are usually contained within the first few documents. 
The suboptimal performance observed on the HotpotQA dataset when $k=1$ can be attributed to the dataset requiring two documents to calculate R-Precision.

\textbf{Impact of Beam Search Sizes.} 
Figures \ref{fig_first_beam_size} and \ref{fig_second_beam_size} show the impact of setting different beam sizes in the first and second stages, respectively, in \method when implemented based on LLaMA-2-13b. 
For the first stage of recalling title identifiers, a larger beam size can achieve better page-level results, thereby slightly improving the effectiveness of the second stage of passage recall. 
However, in the second stage of fine-grained passage recall, the improvement brought by a larger beam size is not significant and may even lead to a slight decline, possibly due to the introduction of additional noise by a larger beam size.

\textbf{Memory Usage Analysis.} Dense retrieval methods such as Contriever and DPR require over 60GB of memory usage. 
In contrast, sparse retrieval methods use far less memory, only needing 17GB. The \method framework, utilizing FM-index and Trie indexing, requires only 8GB when pre-encoding and storing all documents with FM-index, and the Trie storing all title identifiers needs just 25MB, which is negligible. 
Compared to sparse and dense retrieval methods, the recall framework effectively saves memory.

\subsection{Case Study}
\label{subsec:case}

In Supplementary Material Tables \ref{tab_case_study1} to \ref{tab_case_study6}, we present reference cases obtained using the Gold Standard, BM25, and the \method framework with LLaMA-2-13b on the NQ, TriviaQA, and HotpotQA datasets. By generating passages more aligned with the question, \method achieves results that contain the answer in Supplementary Material Tables \ref{tab_case_study1}, \ref{tab_case_study3}, and \ref{tab_case_study5}. Supplementary Material Table \ref{tab_case_study2} showcases a biology question; although the passage recalled and located by \method does not contain the annotated answer, it provides a more detailed description of the location and process of pancreatic enzyme cleavage of peptide bonds. However, Supplementary Material Tables \ref{tab_case_study4} and \ref{tab_case_study6} show instances where \method's recall failed. This is because merely generating a relevant prefix sometimes cannot ensure that the subsequent part will definitely contain the answer, leading to passages that are only broadly related. Ensuring the flexibility of recall and location while considering more subsequent information still requires further exploration. Nevertheless, we can also note that the references obtained by LLM recall are more natural and easier to read compared to those with predefined segmented beginnings. Finally, compared to the NQ dataset, questions in the TriviaQA and HotpotQA datasets are longer, allowing for more vocabulary overlap between questions and passages, which enhances the effectiveness of BM25 sparse retrieval.

\begin{table*}[htbp]
\centering
\begin{tabular}{p{15.5cm}}
\toprule
\textbf{Query:} what is the legal age for marriage in australia   \\
\midrule \textbf{Gold Standard:} Part II (s.10-21) deals with the marriageable age and the marriage of minors. In the original 1961 Act, marriageable age was set at 16 for females and \colorbox{gray!40}{18} for males. However, under section 12 of the original 1961 Act a female 14 or 15 years or a male 16 or 17 years could apply to the court for permission to marry. The marriageable age was equalised in 1991 by the "Sex Discrimination Amendment Act 1991", which raised the marriageable age of females to \colorbox{gray!40}{18}. \\
\midrule \textbf{BM25:} marriage. Before 2004, there was no definition of marriage in the 1961 Act, and instead the common law definition used in the English case "Hyde v Hyde" (1866) was considered supreme. Though s.46(1) of the Act required celebrants to explain the legal nature of marriage in Australia to a couple as "the union of a man and a woman to the exclusion of all others, voluntarily entered into for life", these words were descriptive or explanatory, rather than outlining what constituted a legally valid marriage in \\
\midrule \textbf{\method:} Marriage in Australia is regulated by the federal "Marriage Act 1961" (Cth) , which applies uniformly throughout Australia (including its external territories) to the exclusion of all state laws on the subject. Australian law recognises only monogamous marriages, being marriages of two people, including same-sex marriages, and does not recognise any other forms of union, including traditional Aboriginal marriages, polygamous marriages or concubinage. The marriage age for marriage in Australia is \colorbox{gray!40}{18} years, but in "unusual and exceptional circumstances" a person aged 16 or 17 can marry with parental consent \\
\bottomrule
\end{tabular}
\caption{
In the NQ dataset, the Gold Standard, BM25, and \method reference passage for the query "what is the legal age for marriage in Australia" are provided. The parts containing the answer are highlighted with a \colorbox{gray!40}{grey} background.
}
\label{tab_case_study1}
\end{table*}

\begin{table*}[htbp]
\centering
\begin{tabular}{p{15.5cm}}
\toprule
\textbf{Query:} where does cleavage of the peptide bond by chymotrypsin occur   \\
\midrule \textbf{Gold Standard:} "In vivo", chymotrypsin is a proteolytic enzyme (serine protease) acting in the \colorbox{gray!40}{digestive systems} of many organisms. It facilitates the cleavage of peptide bonds by a hydrolysis reaction, which despite being thermodynamically favorable, occurs extremely slowly in the absence of a catalyst. The main substrates of chymotrypsin are peptide bonds in which the amino acid N-terminal to the bond is a tryptophan, tyrosine, phenylalanine, or leucine. Like many proteases, chymotrypsin also hydrolyses amide bonds "in vitro", a virtue that enabled the use of substrate analogs such as N-acetyl-L-phenylalanine p-nitrophenyl amide for enzyme assays. \\
\midrule \textbf{BM25:} 149, producing $\alpha$-chymotrypsin (which is more active and stable than $\pi$-chymotrypsin). The resulting molecule is a three-polypeptide molecule interconnected via disulfide bonds. "In vivo", chymotrypsin is a proteolytic enzyme (serine protease) acting in the \colorbox{gray!40}{digestive systems} of many organisms. It facilitates the cleavage of peptide bonds by a hydrolysis reaction, which despite being thermodynamically favorable, occurs extremely slowly in the absence of a catalyst. The main substrates of chymotrypsin are peptide bonds in which the amino acid N-terminal \\
\midrule \textbf{\method:} Chymotrypsin cleaves peptide bonds by attacking the unreactive carbonyl group with a powerful nucleophile, the serine 195 residue located in the active site of the enzyme, which briefly becomes covalently bonded to the substrate, forming an enzyme-substrate intermediate. Along with histidine 57 and aspartic acid 102, this serine residue constitutes the catalytic triad of the active site. These findings rely on inhibition assays and the study of the kinetics of cleavage of the aforementioned substrate, exploiting the fact that the en \\
\bottomrule
\end{tabular}
\caption{
In the NQ dataset, the Gold Standard, BM25, and \method reference passage for the query "where does cleavage of the peptide bond by chymotrypsin occur" are provided. The parts containing the answer are highlighted with a \colorbox{gray!40}{grey} background.
}
\label{tab_case_study2}
\end{table*}

\begin{table*}[htbp]
\centering
\begin{tabular}{p{15.5cm}}
\toprule
\textbf{Query:} Which expression is associated with the sinking of the HMS Birkenhead at Gansbaai near Cape Town, South Africa, in Febuary 1852?   \\
\midrule \textbf{Gold Standard:} Only 193 of the estimated 643 people on board survived, and the soldiers' chivalry gave rise to the unofficial "women and children first" protocol when abandoning ship, while the \colorbox{gray!40}{"Birkenhead drill"} of Rudyard Kipling's poem came to describe courage in face of hopeless circumstances. \\
\midrule \textbf{BM25:} HMS "Birkenhead, also referred to as HM Troopship "Birkenhead or Steam Frigate "Birkenhead", was one of the first iron-hulled ships built for the Royal Navy. She was designed as a steam frigate, but was converted to a troopship before being commissioned. She was wrecked on 26 February 1852, while transporting troops to Algoa Bay at Danger Point near Gansbaai, 87 miles (140 kilometres) from Cape Town in the Cape Colony. There were not enough serviceable lifeboats for all the passengers, and the soldiers \\
\midrule \textbf{\method:} The sinking of the "Birkenhead" is one of the earliest maritime disaster evacuations during which the concept of "women and children first" is known to have been applied."Women and children first" subsequently became standard procedure in relation to the evacuation of sinking ships, in fiction and in life. The term \colorbox{gray!40}{"Birkenhead drill"} became defined as courageous behaviour in hopeless circumstances and appeared in Rudyard Kipling's 1893 tribute to the Royal Marines, "Soldier an' Sailor Too": To take your chance in the thick of a rush, with firing all about, Is nothing so bad when you've \\
\bottomrule
\end{tabular}
\caption{
In the TriviaQA dataset, the Gold Standard, BM25, and \method reference passage for the query "Which expression is associated with the sinking of the HMS Birkenhead at Gansbaai near Cape Town, South Africa, in Febuary 1852?" are provided. The parts containing the answer are highlighted with a \colorbox{gray!40}{grey} background.
}
\label{tab_case_study3}
\end{table*}

\begin{table*}[htbp]
\centering
\begin{tabular}{p{15.5cm}}
\toprule
\textbf{Query:} Who was the first winner of `I'm A Celebrity, Get Me Out Of Here'?   \\
\midrule \textbf{Gold Standard:} The first series of I'm a Celebrity...Get Me Out of Here! was broadcast on ITV from 25 August to 8 September 2002. Ant \& Dec presented the main show on ITV, whilst Louise Loughman hosted the spin-off show "I'm a Celebrity...Get Me Out of Here! NOW!" on ITV2. The winner of this series was radio DJ \colorbox{gray!40}{Tony Blackburn}. \\
\midrule \textbf{BM25:} The first series of I'm a Celebrity...Get Me Out of Here! was broadcast on ITV from 25 August to 8 September 2002. Ant \& Dec presented the main show on ITV, whilst Louise Loughman hosted the spin-off show "I'm a Celebrity...Get Me Out of Here! NOW!" on ITV2. The winner of this series was radio DJ \colorbox{gray!40}{Tony Blackburn}. The show began with 8 celebrity contestants. The contestants take part in daily trials to earn food All ratings are taken from the UK \\
\midrule \textbf{\method:} I'm a Celebrity...Get Me Out of Here! is a British reality TV series in which a number of celebrities live together in a jungle environment for a number of weeks, competing to be crowned "King" or "Queen of the Jungle". The show was originally created in the United Kingdom by the factual programmes department of ITV's then London franchise, London Weekend Television and developed by a team including James Allen, Natalka Znak, Brent Baker and Stewart Morris. The first episode aired on 25 August 2002 . It is now produced by ITV Studios and has been licensed globally to countries including the United States \\
\bottomrule
\end{tabular}
\caption{
In the TriviaQA dataset, the Gold Standard, BM25, and \method reference passage for the query "Who was the first winner of `I'm A Celebrity, Get Me Out Of Here'?" are provided. The parts containing the answer are highlighted with a \colorbox{gray!40}{grey} background.
}
\label{tab_case_study4}
\end{table*}

\begin{table*}[htbp]
\centering
\begin{tabular}{p{15.5cm}}
\toprule
\textbf{Query:} 2014 S/S is the debut album of a South Korean boy group that was formed by who?   \\
\midrule \textbf{Gold Standard:} 2014 S/S is the debut album of South Korean group WINNER. It was released on August 12, 2014 by the group's record label, \colorbox{gray!40}{YG Entertainment}. The members were credited for writing the lyrics and composing the majority of the album's songs. \\
\midrule \textbf{BM25:} S is a South Korean project group consisting of three members: Kangta, Lee Ji-hoon and Shin Hye-sung. The group debuted in 2003, under the SM Entertainment label. After 11 years, they released and promoted another mini-album in 2014. In 2003, S released their first album "Fr.In.Cl", which stands for "Friends in Classic". In 2014, after more than a decade in hiatus, the group released their second mini-album "Autumn \\
\midrule \textbf{\method:} 2014 S/S is the debut album of South Korean group WINNER. It was released on August 12, 2014 by the group's record label, \colorbox{gray!40}{YG Entertainment}. The members were credited for writing the lyrics and composing the majority of the album's songs. The members produced the majority of the music for this album themselves, with the help of other producers such as Choice 37, B.I., Airplay, and others. The album was highlighted for incorporating elements generally absent from K-pop releases, including hints of acoustic \\
\bottomrule
\end{tabular}
\caption{
In the HotpotQA dataset, the Gold Standard, BM25, and \method reference passage for the query "2014 S/S is the debut album of a South Korean boy group that was formed by who?" are provided. The parts containing the answer are highlighted with a \colorbox{gray!40}{grey} background.
}
\label{tab_case_study5}
\end{table*}

\begin{table*}[htbp]
\centering
\begin{tabular}{p{15.5cm}}
\toprule
\textbf{Query:} Who is the fictional head of a British Secret Service division and for which a one-time missionary was the inspiration for? \\
\midrule \textbf{Gold Standard:} Charles Fraser-Smith (26 January 1904 – 9 November 1992) was an author and one-time missionary who is widely credited as being the inspiration for Ian Fleming's James Bond quartermaster \colorbox{gray!40}{Q}. During World War II, Fraser-Smith worked for the Ministry of Supply, fabricating equipment nicknamed "\colorbox{gray!40}{Q}-devices" (after \colorbox{gray!40}{Q}-ships) for SOE agents operating in occupied Europe. Prior to the war, Fraser-Smith had worked as a missionary in North Africa. After the war he purchased a dairy farm in Bratton Fleming, Devon, where he died in 1992. \\
\midrule \textbf{BM25:} \colorbox{gray!40}{Q} is a fictional character in the James Bond films and film novelisations. \colorbox{gray!40}{Q} (standing for Quartermaster), like M, is a job title rather than a name. He is the head of \colorbox{gray!40}{Q} Branch (or later \colorbox{gray!40}{Q} Division), the fictional research and development division of the British Secret Service. The use of letters as pseudonyms for senior officers in the British Secret Intelligence Service was started by its first director Captain Sir Mansfield George Smith-Cumming (1859-1923) who signed himself with a C written in \\
\midrule \textbf{\method:} Ian Fleming created the fictional character of James Bond as the central figure for his works. Bond is an intelligence officer in the Secret Intelligence Service, commonly known as MI6. Bond is known by his code number, 007, and was a Royal Naval Reserve Commander. Fleming based his fictional creation on a number of individuals he came across during his time in the Naval Intelligence Division and 30 Assault Unit during the Second World War, admitting that Bond "was a compound of all the secret agents and commando types I met during the war". Among those types were his brother, Peter, who had been involved in behind-the-lines operations in Norway and Greece during the war. \\
\bottomrule
\end{tabular}
\caption{
In the HotpotQA dataset, the Gold Standard, BM25, and \method reference passage for the query "Who is the fictional head of a British Secret Service division and for which a one-time missionary was the inspiration for?" are provided. The parts containing the answer are highlighted with a \colorbox{gray!40}{grey} background.
}
\label{tab_case_study6}
\end{table*}

\end{document}